%% file: main.tex
\def\year{2021}\relax
\documentclass[letterpaper]{article} 
\usepackage{aaai21}  
\usepackage{times}  
\usepackage{helvet} 
\usepackage{courier}  
\usepackage[hyphens]{url}  
\usepackage{graphicx} 
\usepackage{natbib}  
\usepackage{caption} 
\usepackage{subcaption}
\usepackage{graphicx}
\usepackage{amsmath}
\usepackage{makecell}
\usepackage{amssymb}
\usepackage{booktabs}
\usepackage[ruled,vlined]{algorithm2e}
\newtheorem{theorem}{Theorem}[section]
\title{Adversarial Turing Patterns from Cellular Automata}

\title{Adversarial Turing Patterns from Cellular Automata}
\author {
    Nurislam Tursynbek\textsuperscript{\rm 1},
    Ilya Vilkoviskiy\textsuperscript{\rm 1},
    Maria Sindeeva\textsuperscript{\rm 1}, Ivan Oseledets\textsuperscript{\rm 1} \\
}
\affiliations {
    \textsuperscript{\rm 1}Skolkovo Institute of Science and Technology \\
    nurislam.tursynbek@gmail.com,  reminguk@gmail.com,  maria.sindeeva@skoltech.ru,  i.oseledets@skoltech.ru
}
\begin{document}

\maketitle

\begin{abstract}
State-of-the-art deep classifiers are intriguingly vulnerable to universal adversarial perturbations: single disturbances of small magnitude that lead to misclassification of most inputs. This phenomena may potentially result in a serious security problem. Despite the extensive research in this area, there is a lack of theoretical understanding of the structure of these perturbations. In image domain, there is a certain visual similarity between patterns, that represent these perturbations, and classical Turing patterns, which appear as a solution of non-linear partial differential equations and are underlying concept of many processes in nature. In this paper, we provide a theoretical bridge between these two different theories, by mapping a simplified algorithm for crafting universal perturbations to (inhomogeneous) cellular automata, the latter is known to generate Turing patterns. Furthermore, we propose to use Turing patterns, generated by cellular automata, as universal perturbations, and experimentally show that they significantly degrade the performance of deep learning models. We found this method to be a fast and efficient way to create a data-agnostic quasi-imperceptible perturbation in the black-box scenario. The source code is available at https://github.com/NurislamT/advTuring. 
\end{abstract}

\section{Introduction}

Deep neural networks have shown success in solving complex problems for different applications ranging from medical diagnoses to self-driving cars, but recent findings surprisingly show they are not safe and vulnerable to well-designed negligibly perturbed inputs \cite{szegedy2013intriguing, goodfellow2015explaining}, called adversarial examples, compromising people’s confidence in them. Moreover, most of modern defenses to adversarial examples are found to be easily circumvented \cite{athalye2018obfuscated}. One reason why adversarial examples are hard to defend against is the difficulty of constructing a theory of the crafting process of them.

Intriguingly, adversarial perturbations can be transferable across inputs. Universal Adversarial Perturbations (UAPs), single disturbances of small magnitude that lead to misclassification of most inputs, were presented in image domain by Moosavi-Dezfooli et. al \cite{moosavi2017universal}, where authors proposed iterative strategy of gradually pushing a data point to the decision boundary. However, to construct a successful perturbation thousands of images were needed, whereas Khrulkov et. al \cite{khrulkov2018art} proposed an efficient algorithm of constructing UAPs with a very small number of samples. The proposed universal perturbations construct complex and interesting unusual patterns. Studying how these patterns emerge will allow better understanding  the nature of adversarial examples. 
\begin{figure*}[t]
\begin{subfigure}{.64\textwidth}
  \centering
  \includegraphics[width=\textwidth]{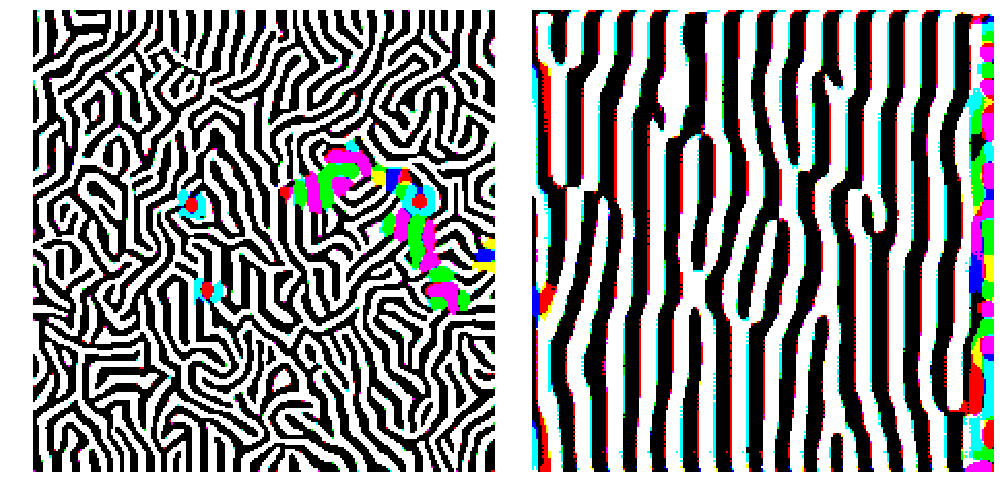}
  \caption{Example of UAPs constructed by \protect\cite{khrulkov2018art}}
  \label{fig:sfig1}
\end{subfigure}%
\begin{subfigure}{.34\textwidth}
  \centering
  \includegraphics[width=\textwidth]{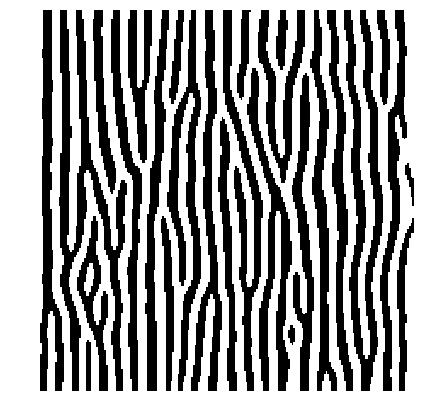}
  \caption{Turing patterns}
  \label{fig:sfig2}
\end{subfigure}
\caption{Visual similarity of Universal Adversarial Perturbations by \protect\cite{khrulkov2018art} and Turing patterns}
\label{fig:fig1}
\end{figure*}

We start from an interesting observation that patterns generated in \cite{khrulkov2018art} visually look very similarly to the so-called \emph{Turing patterns} (Figure \ref{fig:fig1}) which were introduced by Alan Turing in the seminal paper \textit{``The Chemical Basis of Morphogenesis''} \cite{turing1990chemical}. It describes the way in which patterns in nature such as stripes and spots can arise naturally out of a homogeneous uniform state. The original theory of Turing patterns, a two-component reaction-diffusion system, is an important model in mathematical biology and chemistry. Turing found that a stable state in the system with local interactions can become unstable in the presence of diffusion. Reaction–diffusion systems have gained significant attention and was used as a prototype model for pattern formation.

In this paper, we provide an explanation why UAPs from \cite{khrulkov2018art} bear similarity to the Turing patterns using the formalism of \textit{cellular automata} (CA): the iterative process for generating UAPs can be approximated by such process, and Turing patterns can be easily generated by cellular automata \cite{young1984local}. Besides, this gives a very simple way to generate new examples by learning the parameters of such automata by black-box optimization. We also experimentally show this formalism can produce examples very close to so-called single Fourier attacks by studying Fourier harmonics of the obtained examples.

\textbf{The main contributions of the paper are following:}
\begin{itemize}
    \item We show that the iterative process to generate Universal Adversarial Perturbations from \cite{khrulkov2018art} can be reformulated as a cellular automata that generates Turing patterns.
    \item We experimentally show Turing patterns can be used to generate UAPs in a black-box scenario with high fooling rates for different networks.
\end{itemize}

\section{Background}
\subsection{Universal Adversarial Perturbations}
Adversarial perturbations are small disturbances added to the inputs that cause machine learning models to make a mistake. In \cite{szegedy2013intriguing} authors discovered these noises by solving the optimization problem:
\begin{equation}\label{turing:adv-ex}
\min_{\varepsilon}\|\varepsilon\|_2\quad s.t.\;\; \mathcal{C}(\mathbf{x}+\varepsilon)\neq\mathcal{C}(\mathbf{x}),
\end{equation}
where $\mathbf{x}$ is an input object and $\mathcal{C}(\cdot)$ is a neural network classifier. It was shown that the solution to the minimization model \eqref{turing:adv-ex} leads to perturbations imperceptible to human eye. 


Universal adversarial perturbation \cite{moosavi2017universal} is a small  ($\|\varepsilon\|_p\leq L$) noise that makes classifier to misclassify the fraction ($1-\delta$) of inputs from the given dataset $\mu$. 
The goal is to make $\delta$ as small as possible and find a perturbation $\varepsilon$ such that:
\begin{equation}
\mathbb{P}_{\mathbf{x}\sim\mu}\left[\mathcal{C}(\mathbf{x}+\varepsilon)\neq\mathcal{C}(\mathbf{x})\right]\geq 1- \delta\quad s.t.\;\; \|\varepsilon\|_p\leq L,
\end{equation}

In \cite{khrulkov2018art} an efficient way of computing such UAPs was proposed, achieving relatively high fooling rates using only small number of inputs. Consider an input $\mathbf{x}\in\mathbb{R}^d$, its $i$-th layer output $\mathbf{f}_i(\mathbf{x}) \in\mathbb{R}^{d_i}$ and Jacobian matrix $\mathbf{J}_i(x)=\left.\frac{\partial \mathbf{f}_i(\mathbf{x})}{\partial \mathbf{x}}\right|_\mathbf{x}\in\mathbb{R}^{d_i\times d}$. For a small input perturbation $\varepsilon\in\mathbb{R}^d$, using first-order Taylor expansion $\mathbf{f}_i(\mathbf{x}+\varepsilon)\approx \mathbf{f}_i(\mathbf{x})+ \mathbf{J}_i(\mathbf{x})\varepsilon$, authors find that to construct a UAP, it is sufficient to maximize  the sum of norms of Jacobian matrix product with perturbation for a small batch of inputs $X_b$, constrained with perturbation norm ($\|\varepsilon\|_p=L$ is obtained by multiplying the solution by $L$):
\begin{equation}
\sum_{\mathbf{x}_j\in X_b}\|\mathbf{J}_i(\mathbf{x}_j)\varepsilon\|_q^q\rightarrow\max, \qquad \text{s.t. } \|\varepsilon\|_p=1.
\label{sf}
\end{equation}
To solve the optimization problem \eqref{sf} the Boyd iteration \cite{boyd1974power} (generalization of the power method to the problem of computing generalized singular vectors) is found to provide fast convergence:
\begin{equation}\label{turing:boyd}
\varepsilon_{t+1}=\frac{\psi_{p'}(\mathbf{J}_i^{T}(X_b)\psi_q(\mathbf{J}_i(X_b)\varepsilon_t))}{\|\psi_{p'}(\mathbf{J}_i^{T}(X_b)\psi_q(\mathbf{J}_i(X_b)\varepsilon_t))\|_p},
\end{equation}
where $\frac{1}{p'}+\frac{1}{p}=1$ and $\psi_r(z)=\mathrm{sign}(z)|z|^{r-1}$, and $\mathbf{J}_i(X_b)\in\mathbb{R}^{bd_i\times d}$ for a batch $X_b$ with batch size $b$ is given as a block matrix:
\begin{equation}
\mathbf{J}_i(X_b) = \begin{bmatrix}
\mathbf{J}_i(\mathbf{x}_1)\\
\vdots\\
\mathbf{J}_i(\mathbf{x}_b)
\end{bmatrix}.
\end{equation}
For the case of $p=\infty$ \eqref{turing:boyd} takes the form:
\begin{equation}\label{Boyd}
\varepsilon_{t+1}=\mathrm{sign}(\mathbf{J}_i^{T}(X_b)\psi_q(\mathbf{J}_i(X_b)\varepsilon_t)).
\end{equation}
Equation \eqref{Boyd} is the first crucial point in our study, and we will show, how they connect to Turing patterns. We now proceed with describing the background behind these patterns and mathematical correspondence between Equation \eqref{Boyd} and emergence of Turing patterns as cellular automata is described in next Section.


\subsection{Turing Patterns as Cellular Automata}
\label{tuca}
In his seminal work \cite{turing1990chemical} Alan Turing studied the emergence theory of patterns in nature such as stripes and spots that can arise out of a homogeneous uniform state. The proposed model was based on the solution of reaction-diffusion equations of two chemical morphogens (reagents):
\begin{equation}
\begin{split}
\frac{\partial n_1(i,j)}{\partial t}=-\mu_1 \nabla^2 n_1(i,j)+ a(n_1(i,j),n_2(i,j)), \\
\frac{\partial n_2(i,j)}{\partial t}=-\mu_2 \nabla^2 n_2(i,j)+ b(n_1(i,j),n_2(i,j)).
\end{split}
\label{RD1}
\end{equation}

Here, $n_1(i,j)$ and $n_2(i,j)$ are concentrations of two morphogens in the point with coordinates $(i,j)$ $\mu_1$ and $\mu_2$ are scalar coefficients. $a$ and $b$ are nonlinear functions, with at least two points $(i,j)$, satisfying $(i,j):\begin{cases} a(n_1(i,j),n_2(i,j))=0 \\ b(n_1(i,j),n_2(i,j))=0. \end{cases}$

Turing noted the solution presents alternating patterns with specific size that does not depend on the coordinate and describes the stationary solution, which interpolates between zeros of $a$ and $b$.

Young et. al \cite{young1984local} proposed to generate Turing patterns by a discrete cellular automata as following. Let us consider a 2D grid of cells. Each cell $(i,j)$ is equipped with a number $n(i,j)\in\{0,1\}$. The sum of cells, neighbouring with the current cell $(i,j)$ within the radius $r_i$ is multiplied by $w$, while the sum of values of cells, neighbouring with the current cell $(i,j)$ between radii $r_{in}$ and $r_{out}$, is multiplied by $-1$. If the total sum of these two terms is positive, the new value of the cell is $1$, otherwise $0$. 
This process can be written by introducing the convolutional kernel $Y(m,l)$:
\begin{equation}
Y(m,l)=\begin{cases}
w & \text{if }  |m|^2+|l|^2 <r_{in}^2, \\
-1 & \text{if }  r^2_{out}>|m|^2+|l|^2 >r_{in}^2.
\end{cases}
\end{equation}
The coefficient $w$ is found from the condition that the sum of elements of kernel $Y$ is set to be $0$:
\begin{equation}
\sum\limits_{m}\sum\limits_{l}Y(m,l)=0. \label{balance}
\end{equation}
The update rule is written as:
\begin{align}
n_{k+1}(i,j)=\frac{1}{2}\left[\mathrm{sign}\left(Y\star n_k(i,j)\right)+1\right] \label{step},
\end{align}
where $\star$ denotes a usual 2D convolution operator:
\begin{equation}
Y\star n(i,j)=\sum\limits_{m}\sum\limits_{l}Y(m,l)n(i+m,j+l).
\end{equation}
For convenience, instead of $\{0,1\}$, we can use $\pm 1$ and rewrite (\ref{step}) (considering condition (\ref{balance})) as:
\begin{equation}
    n_{k+1}(i,j)=\mathrm{sign}\Big(Y\star n_k(i,j)\Big).
    \label{sign}
\end{equation}
Note that \eqref{sign} is similar to \eqref{Boyd}, and lets investigate this connection in more details.

\section{Connection between Boyd iteration and Cellular Automata}
\label{connection}


We restrict our study to the case $q=2$. Then $\psi_2(z)=z$ and Equation (\ref{Boyd}) can be rewritten as:
\begin{equation} \label{Boyd_cell}
\varepsilon_{t+1}=\mathrm{sign}(\mathbf{J}_i^{T}(X_b)\mathbf{J}_i(X_b) \varepsilon_t).
\end{equation}

In \cite{khrulkov2018art} it was shown that perturbations of the first layers give the most promising results. Initial layers of most of the state-of-the-art image classification neural networks are convolutional followed by nonlinear activation functions $\sigma$ (e.g. $\text{ReLU}(z)=\max(0,z)$). If the network does not include other operations (such as pooling or skip connections) the output of the $i$-th layer is:
\begin{equation}
\mathbf{f}_i(\mathbf{x}) = 
\sigma(\mathbf{M}_i \mathbf{f}_{i-1}(\mathbf{x})),
\end{equation}
where $\mathbf{M}_i\in\mathbb{R}^{d_i\times d_{i-1}}$ is a matrix of the 2D convolution operator of the $i$-th layer, corresponding to the convolutional kernel $\mathbf{K}_i$.
Using chain rule, the Jacobian in \eqref{Boyd} is:
\begin{equation}
\mathbf{J}_i(\mathbf{x})=\frac{\partial \mathbf{f}_i(\mathbf{x})}{\partial \mathbf{f}_{i-1}(\mathbf{x})}\cdots\frac{\partial \mathbf{f}_1(\mathbf{x})}{\partial \mathbf{x}} =\mathbf{D}_i(\mathbf{x}) \mathbf{M}_i\cdots \mathbf{D}_1(\mathbf{x}) \mathbf{M}_1, \label{defect}
\end{equation}
where $\mathbf{D}_j(\mathbf{x})=\mathrm{diag}(\theta(\mathbf{M}_j\mathbf{f}_{j-1}(\mathbf{x})))\in\mathbb{R}^{d_j\times d_j}$ and $\theta(z)=\frac{\partial \text{ ReLU}(z)}{\partial z}=\begin{cases} 1,  \text{if } z>0\\ 0,  \text{if } z<0, \end{cases}.$

The update matrix from Equation \eqref{Boyd_cell} is then:
\begin{multline}
\mathbf{J}_i^{T}(X_b)\mathbf{J}_i(X_b)= \left[
\mathbf{J}_i^T(\mathbf{x}_1) \cdots \mathbf{J}_i^T(\mathbf{x}_b)
\right]
\begin{bmatrix}
\mathbf{J}_i(\mathbf{x}_1)\\
\vdots\\
\mathbf{J}_i(\mathbf{x}_b)
\end{bmatrix} =\\ =\sum_{\mathbf{x}\in X_b}\mathbf{J}_i^{T}(\mathbf{x})\mathbf{J}_i(\mathbf{x}).
\label{short}
\end{multline}
Performance of the UAPs from \cite{khrulkov2018art} increases with the increase of batch size. Then considering the limit case of sufficiently large batch size $b$ we can approximate the averaging in \eqref{short} by the expected value:
\begin{equation}
\begin{gathered}
\frac{\sum\limits_{\mathbf{x}\in X_b}\mathbf{J}_i^{T}(\mathbf{x})\mathbf{J}_i(\mathbf{x})}{b}\approx \mathbb{E}_{\mathbf{x}}\left[\mathbf{J}_i^{T}(\mathbf{x})\mathbf{J}_i(\mathbf{x})\right]=\\
=\mathbb{E}\left[\mathbf{M}_1^T\mathbf{D}_1^T(\mathbf{x})\cdots \mathbf{M}_i^T\mathbf{D}_i^T(\mathbf{x}) \mathbf{D}_i(\mathbf{x}) \mathbf{M}_i\cdots \mathbf{D}_1(\mathbf{x}) \mathbf{M}_1\right].
\end{gathered}
    \label{long}
\end{equation}
Note that $\mathbf{D}^T_j(\mathbf{x})\mathbf{D}_j(\mathbf{x})=\mathbf{D}^2_j(\mathbf{x})=\mathbf{D}^T_j(\mathbf{x})=\mathbf{D}_j(\mathbf{x})$.
To simplify Equation \eqref{long}, we make additional assumption.

\textit{Assumption 0.} 
The elements of the diagonal matrices $D_j$ have the same scalar mean value $c_j$:
\begin{equation}
\label{eye}
\mathbb{E}_{\mathbf{x}}\left[\mathbf{D}_j(\mathbf{x})\right]\approx c_j\mathbf{I}.
\end{equation}

This is based on the fact that we consider universal attack, which is input-independent, we can assume $\mathbf{x}$ as random variable and $\mathbf{D}(\mathbf{x})$ as random matrix, which is diagonal and have elements $0$ or $1$. Then our assumption is not unrealistic, as we assume that all diagonal elements of random diagonal matrix over expectation converge to the same number $c$ between $0$ and $1$ (see Appendix). Then, taking into consideration \eqref{short}, \eqref{long}, \eqref{eye} for the first layer:
\begin{multline}
\mathbf{J}_1^{T}(X_b)\mathbf{J}_1(X_b) \approx b\mathbb{E}_{\mathbf{x}}\left[\mathbf{M}_1^T\mathbf{D}_1^T(\mathbf{x})\mathbf{D}_1(\mathbf{x})\mathbf{M}_1\right]=\\=b\mathbf{M}_1^T\mathbb{E}_{\mathbf{x}}\left[\mathbf{D}_1(\mathbf{x})\right]\mathbf{M}_1 =  bc_1\mathbf{M}_1^T\mathbf{M}_1,
\end{multline}
i.e. taking the expected value has removed the term corresponding to the non-linearity! 

Two subsequent convolutions does not result into one convolution, however the only source of error is boundary effects. Since the size of convolutional kernels $d<10$ is much smaller than the dimension of image $N=224$ (for Imagenet), the error is small and is up to $d/N\approx 0.5* 10^{-2}$. It definetely could be neglected for our purposes. Moreover, as shown in \cite{miyato2018spectral}, this is common in convolutional neural networks for images. One can advocate this by the asymptotic theory of Toeplitz matrices \cite{bottcher2000toeplitz}. Thus, we have shown that $\mathbf{J}_1^{T}(X_b)\mathbf{J}_1(X_b)$ has an approximate convolutional structure. To show that $\mathbf{J}_i^{T}(X_b)\mathbf{J}_i(X_b)$ is also convolutional, we introduce additional assumptions:

\begin{figure*}[t]
  \includegraphics[width=\textwidth]{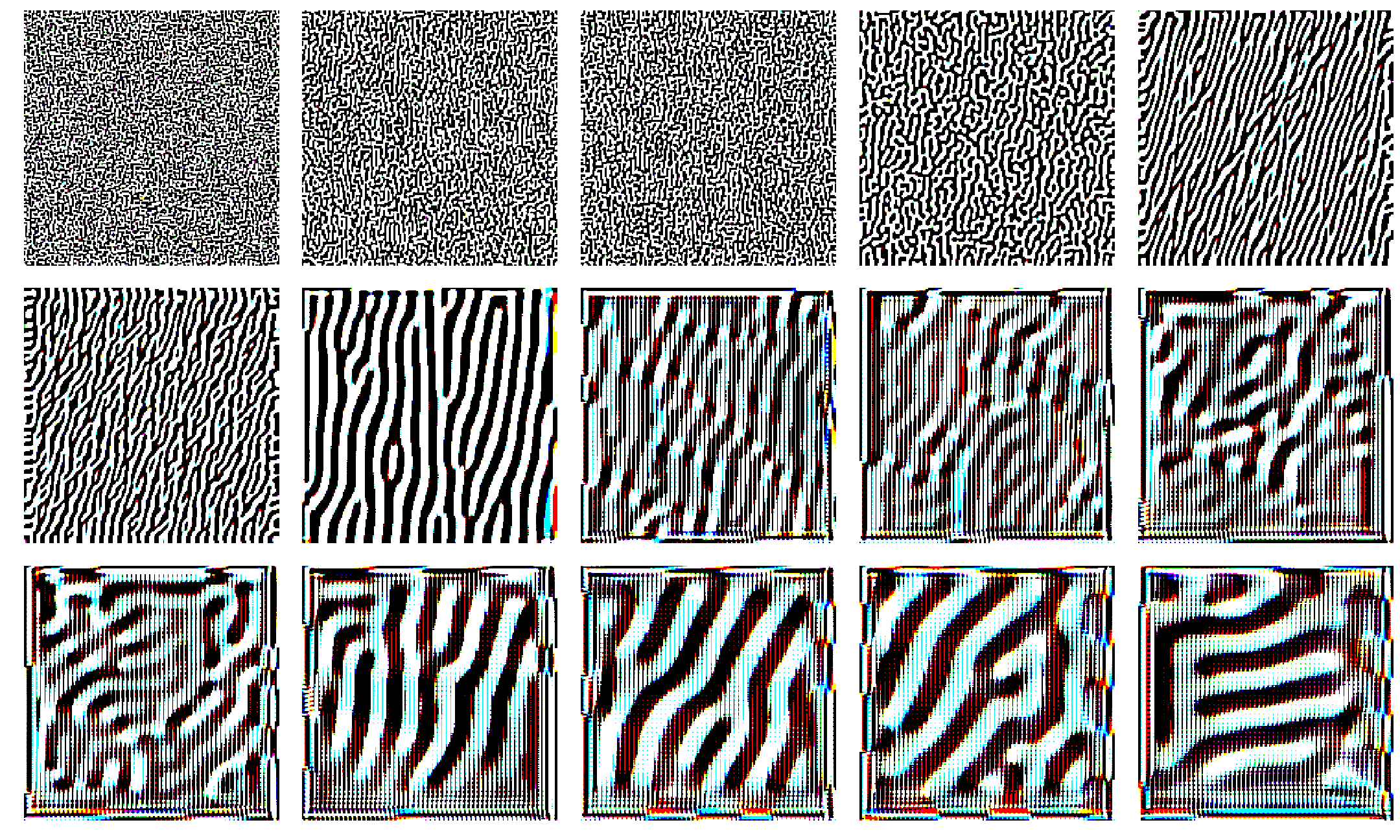}
  \caption{UAPs for different layers of VGG-19 classifier. According to the theory, the specific feature size of the patterns increases with the depth of the attacked layer. The numbering of images is row-wise. First row corresponds to Layers 1-5, second row - Layers 6-10, third row - Layers 11-15.}
  \label{4}
\end{figure*}

\textit{Assumption 1.} Matrices $\mathbf{D}_j(\mathbf{x})$ and $\mathbf{D}_k(\mathbf{x})$ are independent for all pairs of $j$ and $k$. In practice we only will check the uncorrelation instead of independency, these checks are reported in appendix.

\textit{Assumption 2.} Diagonal elements $\mathbf{d}_j=\theta(\mathbf{M}_j\mathbf{f}_{j-1}(\mathbf{x}))$ of matrix $\mathbf{D}_j(\mathbf{x})$ have covariance matrix of convolutional structure $\mathbf{C}_j$ as linear operators of typical convolutional layers:
\begin{equation}
\mathbf{Cov}(\mathbf{d}_j,\mathbf{d}_j) = \mathbf{C}_j.
\end{equation}
These two assumptions might seem to be strong, so here we discuss how realistic and legit they are. 

Regarding assumption 1, we should point that matrix $\textbf{D}$ is not the outputs of ReLU activation function, but is indicator ($0$ and $1$) whether the feature is positive or negative, and this is reasonable to assume the independence between elements $\textbf{D}_i$ and $\textbf{D}_j$. 
Further, let us consider covariaance of matrix elements of diagonal matrices $\mathbf{Cov}(\textbf{D}_{i} , \textbf{D}_{j})$. First of all, for a large enough image batch it is natural to assume that this covariance if translationally invariant, i.e there is no selected points. This means $\mathbf{Cov}(\textbf{D}_{i}^{p,q} , \textbf{D}^{l,m}_{i})=C_i(p-l,q-m)$ which is nothing but the assumption 2. It is also follows that because of finite receptive field, convolution matrix $C_i$ is decreasing far from the diagonal.

These assumptions need additional conditions for satisfactory boundary effects, however we find them natural, and realistic to hold in the main part of features. Additional quantitative experimental justifications of these assumptions are in the Appendix.

\begin{theorem}
\label{pythagorean}
Given $\mathbf{x}$ is a random variable, $\mathbf{A}=\mathbf{A}(\mathbf{x})$ is a square matrix, such that $\mathbb{E}_{x}\left[\mathbf{A}\right]=\mathbf{B}$ is convolutional,  and $\mathbf{D}=\mathbf{D}(\mathbf{x})$ is a diagonal matrix, such that $\mathbb{E}_{\mathbf{x}}\left[\mathbf{D}_{ll} \mathbf{D}_{mm}\right]=\mathbf{C}_{lm}$ and matrices $\mathbf{D}$ and $\mathbf{A}$ are independent. Then, $\mathbb{E}_x\left[\mathbf{D}\mathbf{A}\mathbf{D}\right]$ is a convolutional matrix.
\end{theorem}
\textit{Proof:}
\begin{multline*}
\mathbb{E}_x\left[\mathbf{D}\mathbf{A}\mathbf{D}(\mathbf{x})\right]_{lm}=\mathbb{E}_x\left[\mathbf{D}_{ll}\mathbf{A}_{lm}\mathbf{D}_{mm}\right]=\\=\mathbb{E}_x\left[\mathbf{A}_{lm}\right]\mathbb{E}_x\left[\mathbf{D}_{ll}\mathbf{D}_{mm}\right]=\mathbf{B}_{lm}\mathbf{C}_{lm}=(\mathbf{B}\circ\mathbf{C})_{lm},
\end{multline*}
where $\circ$ denotes element-wise product.

Applying Theorem \ref{pythagorean} for each of the layers, and using $\mathbf{D}_j=\mathbf{D}_j(\mathbf{x})$ for all $j$ in \eqref{short} and \eqref{long}:
\begin{equation}
    \begin{split}
&\mathbf{J}_i^{T}(X_b)\mathbf{J}_i(X_b) =\\&=b \mathbb{E}\left[\mathbf{M}_1^T\mathbf{D}_1^T\overbrace{ \cdots\mathbf{M}_i^T\mathbf{D}_i^T \mathbf{D}_i\mathbf{M}_i\cdots}^{\text{no correlation with }\mathbf{D}_1} \mathbf{D}_1 \mathbf{M}_1\right]=\\
&=b\mathbf{M}_i^T((\cdots\mathbf{M}_2^T((\mathbf{M}_1^T\mathbf{C}_1\mathbf{M}_1)\circ\mathbf{C}_2)\mathbf{M}_2\cdots)\circ\mathbf{C}_i)\mathbf{M}_i,
    \end{split}
\label{jj}
\end{equation}
which concludes that matrix $\mathbf{J}_i^{T}(X_b)\mathbf{J}_i(X_b)$ is convolutional matrix and Boyd algorithm \eqref{Boyd} indeed generates cellular automata \eqref{sign}.

\begin{figure*}[t]
    \centering
    \begin{subfigure}{.33\textwidth}
      \centering
      \includegraphics[width=\textwidth]{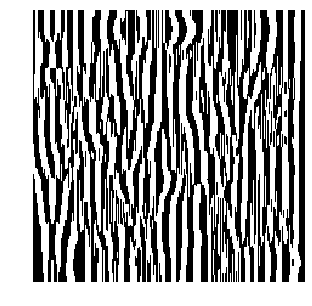}
      \caption{MobileNetV2}
    \end{subfigure}\hfill
    \begin{subfigure}{.33\textwidth}
      \centering
      \includegraphics[width=\textwidth]{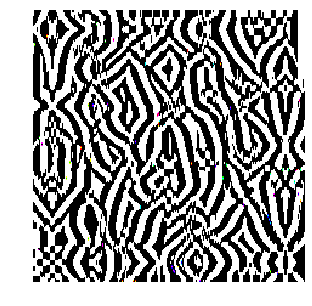}
      \caption{VGG-19}
    \end{subfigure}\hfill
    \begin{subfigure}{.33\textwidth}
      \centering
      \includegraphics[width=\textwidth]{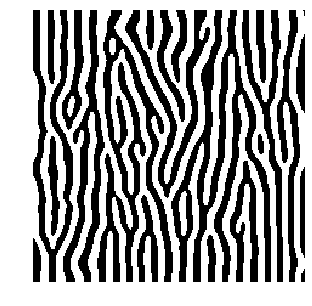}
      \caption{InceptionV3}
    \end{subfigure}
\caption{Turing patterns generated using simplest kernels, parameterized by two numbers $l1,l2$.}
\label{l1l2}
\end{figure*}

\begin{table}[h!]
      \centering
        \begin{tabular}{|l	||c	|c	|c	|c	|}\hline
    	\textbf{Target}	&	\textbf{Random}	&	\textbf{Optimized}	\\	\hline\hline
    	MobileNetV2	&	52.79	&	56.15	\\	\hline
    	VGG-19	&	52.56	&	55.4	\\	\hline
    	InceptionV3	&	45.32	&	47.18	\\	\hline
    \end{tabular}				
    \caption{FRs of a random Turing pattern and the one optimized in $l_1, l_2$ and initialization. Even random Turing patterns provide competitive results, proving high universality of this attack. Better optimizations are in the tables \ref{init},\ref{cannelmix}.}
    \label{tab:params}
        
\end{table}

The structure of patterns by \cite{khrulkov2018art} contains some interesting elements, which has specific size. We empirically find that changing the train batch does not affect much neither form of pattern, nor its fooling rate (Equation \eqref{frformula}), even if we use batch of white images. This supports the idea that the update matrix $\mathbf{J}_i^{T}(X_b)\mathbf{J}_i(X_b)$ does not depend much on batch $X_b$, and can be replaced with constant convolutional matrix by Equation \eqref{jj}. Based on Equation \eqref{jj}, we see that deeper the layer $i$ we apply algorithm in Equation \eqref{Boyd} the more convolutions are involved. As it was mentioned in Section \ref{tuca} Turing patterns present alternating patterns with specific characteristic size. Based on this intuition, we hypothesize that the specific size of the patterns generated by \cite{khrulkov2018art} increases with the depth of the attacked layer for which UAP is constructed. To ensure this, we consider VGG-19 \cite{simonyan2014very} network. Initial layers of this network are convolutional with ReLU activation.  We apply algorithm to each of the first $15$ layers of the network. We chose $q=2$ in Equation \eqref{Boyd}.
As illustrated in Figure \ref{4}, the typical feature size of the patterns increases with the ``depth of the attack''.

\begin{table}[h!]
\centering
\begin{tabular}{l|c|c|c|c}
\toprule
\textbf{Target model} &  \multicolumn{2}{c}{\textbf{With}} & \multicolumn{2}{c}{\textbf{Without}}\\
\midrule
            & \textbf{250 q} & \textbf{500 q} & \textbf{250 q} & \textbf{500 q} \\
MobileNetV2 & 85.62 & 90.91 & 91.67 & 94.47 \\ 
VGG-19      & 79.73 & 75.9  & 76.94 & 77.61 \\ 
InceptionV3 & 52.9  & 54.2  & 53.71 & 53.23 \\ 
\bottomrule
\end{tabular}
\caption{FRs of patterns with and without optimization on initialization (250 and 500 queries)}
\label{initl1l2}
\end{table}

\section{Turing Patterns as Black-box Universal Adversarial Perturbations}
\label{sec4}

\begin{figure*}[t]
      \centering
    \begin{subfigure}{.33\textwidth}
      \centering
      \includegraphics[width=\textwidth]{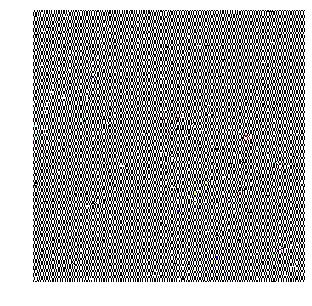}
      \caption{MobileNetV2}
      \end{subfigure}\hfill
    \begin{subfigure}{.33\textwidth}
      \centering
      \includegraphics[width=\textwidth]{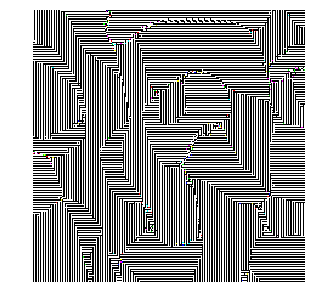}
      \caption{VGG-19}
      \end{subfigure}\hfill
    \begin{subfigure}{.33\textwidth}
      \centering
      \includegraphics[width=\textwidth]{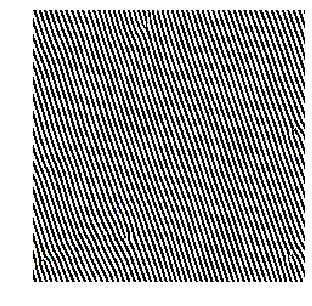}
      \caption{InceptionV3}
      \end{subfigure}
    \caption{Turing patterns generated using optimization on initialization and convolutional kernel $Y$}
    \label{fig:turing_examples}
\end{figure*}

The connection in the previous section suggests the use of Turing patterns to craft UAPs in a black-box setting, learning parameters of CA by directly maximizing the fooling rate.

As described earlier, complex Turing patterns can be modeled with only several parameters. In order to test the performance of Turing patterns as black-box UAPs for a specific network, we learn this parameters in a black-box scenario by maximizing the \textit{fooling rate}, without accessing the model's architecture. For a classifier $f$ and $N$ images, the fooling rate of Turing pattern $\varepsilon(\theta)$ with parameters $\theta$ is:
\begin{equation}
\label{frformula}
\mathrm{FR} = \frac{\sum\limits_{i=1}^N [{\arg\max f(\mathbf{x}_i) \neq\arg\max f(\mathbf{x}_i+\varepsilon(\theta))]}}{N}.
\end{equation}

There are several options how to parametrize CA, and we discuss them one-by-one. For all approaches the target function is optimized on $512$ random images from ImageNet \cite{russakovsky2015imagenet} train dataset, and perturbation is constrained as $\|\varepsilon\|_{\infty}\leq10$. Following \cite{meunier2019yet}, we used evolutionary algorithm CMA-ES \cite{hansen2003reducing} by Nevergrad \cite{rapin2018nevergrad} as a black-box optimization algorithm. Fooling rates are calculated for $10000$ Imagenet validation images for torchvision pretrained models(VGG-19 \cite{simonyan2014very}, InceptionV3 \cite{szegedy2016rethinking}, MobilenetV2 \cite{sandler2018mobilenetv2}).

\begin{table}[t]
    \label{tab:transfer}
        \centering
        \begin{tabular}{|l|c|c|c|c|}\hline
        \diaghead{\theadfont Diag Mixinggg}%
        {Mixing}{Threshold} & \thead{$max$} & \thead{$max - 1$} & \thead{$0.9 * max$} \\
        \hline
        Summation & 77.5 & 78.0 & 77.9 \\
        \hline
        Pointwise & 78.0 & 77.7 & 78.0\\
        \hline
        \end{tabular}
        \caption{Experiments with DFT frequencies for VGG-19}
    \label{tab2}
\end{table}





\textbf{Simple CA.}
Here, we fix the kernel $Y$ in Equation \eqref{sign} to be $L \times L$ (we find that $L=13$ produces best results), with elements filled by $-1$, except for the inner central rectangle of size $l_1\times l_2$ with constant elements such that the sum of all elements of  $Y$ is $0$. Besides $l_1$ and $l_2$, the initialization of $n(i, j)$ from Eq. \eqref{sign} is added as parameters. To reduce black-box queries, initialization is chosen to be $7 \times 7$ square tiles (size $32\times32$) \cite{meunier2019yet} for each of the $3$ maps representing each image channel. $\theta = (l_1,l_2, \texttt{initialization})$.  Resulting UAPs are shown in Fig. \ref{l1l2}. Improvement over the random Turing pattern (with random, not optimized parameters) is shown in Table \ref{tab:params} and full results are in Table \ref{tul1l2}. We should point, here, our method is black-box, i.e. without access to the architecture and weights, and we do not compare it to \cite{khrulkov2018art} as they consider white-box scenario with full access to the network. Moreover, even if we consider black-box scenario, our results using more advanced optimization are significantly better in terms of fooling rate (see Tables 6-7) even comparing to the white-box.

\begin{figure}[t]
\centering
\begin{subfigure}{.23\textwidth}
  \centering
  \includegraphics[width=\textwidth]{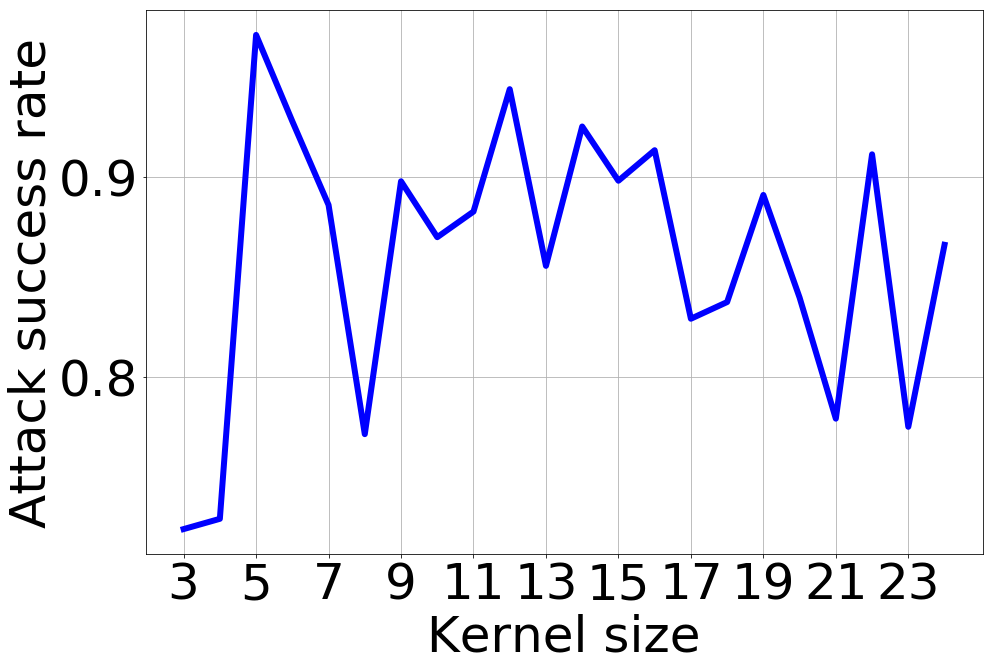}
  \caption{Optimization over kernel and initialization maps}
  \label{5a}
\end{subfigure}%
\hspace{0.02cm}
\begin{subfigure}{.23\textwidth}
  \centering
  \includegraphics[width=\textwidth]{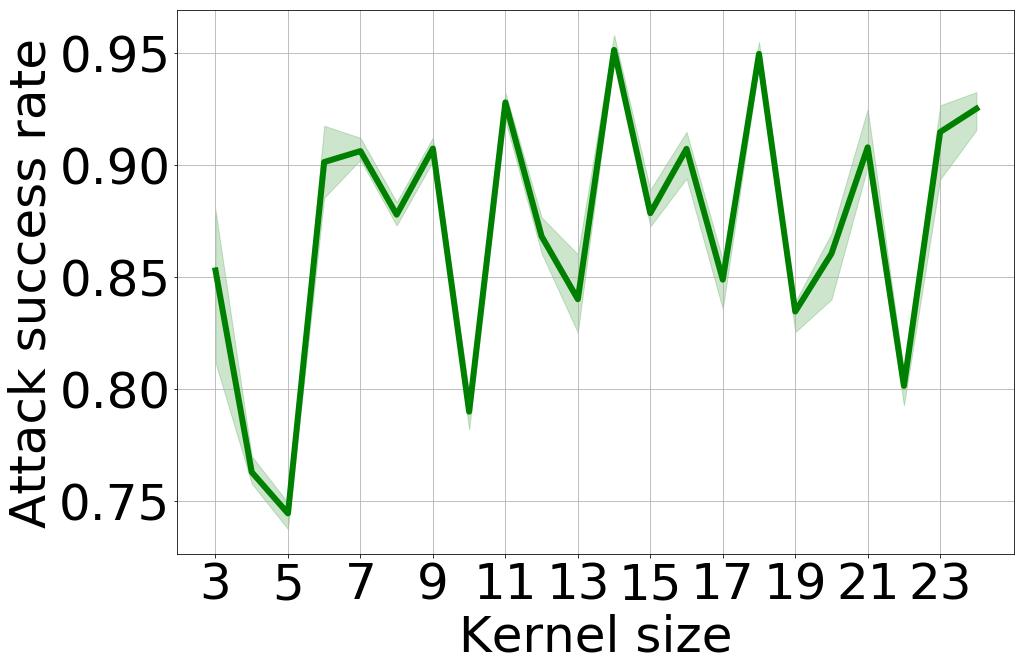}
  \caption{Optimization over kernel only, averaged over 10 initializations (min and max highlighted)}
  \label{5b}
\end{subfigure}
\caption{Pattern fooling rate (vertical axis) dependence on the kernel size (horizontal axis). This figure shows that there is no dependence on the kernel size.}
\label{fig:ranging_filter_size}
\end{figure}

\textbf{CA with kernel optimization.}
In this scenario, all elements of the kernel $Y$ are considered as unknown parameters. Optimizing both over $Y$ and the initialization maps we got a substantial increase in fooling rates (see Fig. \ref{fig:turing_examples} for perturbations and Table \ref{init} for fooling rate results). Results for different kernel sizes are shown in Fig. \ref{5a}.

\textbf{Optimization with random initialization map.}
Here, we remove the initialization map from the optimized parameters select it randomly. As experiments show (see Table \ref{initl1l2}), the performance of patterns generated without optimized initialization maps does not differ significantly from the pattern with optimized initialization maps. Thus, the initialization maps for pattern generation can be randomly initialized, and less queries can be made without significant loss in fooling rates. Results for different kernel sizes are shown in Fig. \ref{5b}.

Another resource for optimization are different strategies for dealing with the channels of the image (This might produce color patches). The detailed experiments are given in the Appendix, the results for the best case are in Table \ref{cannelmix}.

\begin{table}[t]
    \label{tab:transfer}
        \centering
        \begin{tabular}{|l|c|c|c|c|}\hline
        \diaghead{\theadfont Diag Mixinggg}%
        {Mixing}{Threshold} & \thead{$max$} & \thead{$max - 1$} & \thead{$0.9 * max$} \\
        \hline
        Summation & 53.1 & 50.6 & 51.0 \\
        \hline
        Pointwise & 53.3 & 53.6 & 53.5 \\
        \hline
        \end{tabular}
    \caption{Experiments with DFT frequencies for InceptionV3}
     \label{tab3}
\end{table}

\begin{table}[t]
    \label{tab:transfer}
        \centering
        \begin{tabular}{|l|c|c|c|c|}\hline
        \diaghead{\theadfont Diag Trained onn}%
        {Trained on}{Target} & \thead{MobileNetV2} & \thead{VGG-19} & \thead{InceptionV3} \\
        \hline
        MobileNetV2	&	56.15	&	56.96	&	45.16	\\	\hline
        VGG-19	&	50.49	&	55.4	&	49.98	\\	\hline
        InceptionV3	&	49.55	&	48.76	&	47.18	\\	\hline
        \end{tabular}
        \caption{Optimization on $l_1, l_2$, and initialization}
        \label{tul1l2}
\end{table}
\begin{table}[h!]
        \centering
        \begin{tabular}{|l|c|c|c|c|}\hline
        \diaghead{\theadfont Diag Trained onn}%
        {Trained on}{Target} & \thead{MobileNetV2} & \thead{VGG-19} & \thead{InceptionV3} \\
        \hline
        MobileNetV2	&	90.59	&	61.27	&	35.15	\\	\hline
        VGG-19	&	62.15	&	75.64	&	31.91	\\	\hline
        InceptionV3	&	65.69	&	76.40	&	53.93	\\	\hline
        \end{tabular}
        \caption{Optimization on kernel $Y$ and initialization}
        \label{init}
\end{table}
\begin{table}[h!]
      \centering
        \begin{tabular}{|l|c|c|c|c|}\hline
        \diaghead{\theadfont Diag Trained onn}%
        {Trained on}{Target} & \thead{MobileNetV2} & \thead{VGG-19} & \thead{InceptionV3} \\
        \hline
        MobileNetV2	&	94.78	&	57.90	&	35.04	\\	\hline
        VGG-19	&	73.15	&	77.50	&	43.06	\\	\hline
        InceptionV3	&	63.90	&	64.46	&	53.10	\\	\hline
        \end{tabular}
    \caption{Summation channel mixing}
    \label{cannelmix}
\end{table}

\subsection{Similarity to Fourier attacks} \label{SFA}
Some of the patterns are similar to the ones generated as single Fourier harmonics, in \cite{tsuzuku2019structural}. To compare correspondence, we took the discrete Fourier transform (DFT) of the pattern and cut all the frequencies with the amplitude below some threshold (we used $2$ cases of thresholds: $\max - 1$ and $0.9 \max$) and then reversed the DFT for all three pattern channels. We applied this transformation to the patterns acquired using different channel mixing approaches (2 (summation) and 4 (pointwise) described above). As a result, in most cases neither the fooling rates (see Table \ref{tab2}, \ref{tab3}) nor the visual representation of the pattern (Appendix, Fig. 6) did not change drastically, which leads us to the conclusion that in these cases it is indeed a Single Fourier Attack that is taking place. However, in the case of patterns generated using summation channel mixing to attack InceptionV3 it is not true (Appendix, Fig. 7). This means that while such methods can generate Single Fourier Attacks, not all the attacks generated using these methods are SFA, i.e. our approach can generate more diverse UAPs. Our work compared to \cite{tsuzuku2019structural} has much less parameters and performs better (see Table \ref{table8}).
\begin{table}[t]
    \centering
    \begin{tabular}{|c|c|c|c|}
    \hline
          & ResNet  & GoogleNet & VGG-16\\
          \hline
         SFA & $40.1$ & $44.1$ & $53.3$ \\
         Ours & $\textbf{61.4}$ & $\textbf{58.2}$ & $\textbf{81.2}$
         \\\hline
    \end{tabular}
    \caption{Performance of Turing Patterns (2 parameters) and SFA: Single Fourier Attack \cite{tsuzuku2019structural} (224*224*3 parameters) in a black-box scenario}
        \label{table8}
\end{table}


\section{Related Work}

\textbf{Adversarial Perturbations.}
Intriguing vulnerability of neural networks in \cite{szegedy2013intriguing,biggio2013evasion} proposed numerous techniques to construct white-box \cite{goodfellow2015explaining,carlini2017towards,moosavi2016deepfool,madry2017towards} and black-box \cite{papernot2017practical,brendel2017decision,chen2017trainable,ilyas2018black} adversarial examples. Modern countermeasures against adversarial examples have been shown to be brittle \cite{athalye2018obfuscated}. Interestingly, input-agnostic Universal Adversarial Perturbations were discovered recently \cite{moosavi2017universal}. Several methods were proposed to craft UAPs \cite{khrulkov2018art,mopuri2017fast,tsuzuku2019structural}. Mopuri et al. \cite{mopuri2017fast} used activation-maximization approach. Khrulkov et al. \cite{khrulkov2018art} proposed efficient use of $(p,q)-$singular vectors to perturb hidden layers. Tsuzuku et al. \cite{tsuzuku2019structural} used Fourier harmonics of convolutional layers without any activation function. Though UAPs were first found in image classification, they exist in other areas as well, such as semantic segmentation \cite{metzen2017universal}, text classifiers \cite{behjati2019universal}, speech and audio recognition \cite{abdoli2019universal,neekhara2019universal}.
    
\textbf{Turing Patterns.} Initially introduced by Alan Turing in \cite{turing1990chemical}, Turing patterns have attracted a lot of attention in pattern formation studies. Patterns like fronts, hexagons, spirals and stripes are found as a result of Turing reaction–diffusion equations \cite{kondo2010reaction}. In nature, Turing patterns has served as a model of fish skin pigmentation \cite{nakamasu2009interactions}, leopard skin patterns \cite{liu2006two}, seashell patterns \cite{fowler1992modeling}. Interesting to Neuroscience community, might be connection of Turing patterns to brain \cite{cartwright2002labyrinthine,lefevre2010reaction}.

\textbf{Cellular Automata in Deep Learning.} Relations between Neural Networks and Cellular Automata were discovered in several works \cite{wulff1993learning,gilpin2019cellular,mordvintsev2020growing}. Cellular automaton dynamics with neural network was studied in \cite{wulff1993learning}. In \cite{gilpin2019cellular} authors generate cellular automata as convolutional neural network. In \cite{mordvintsev2020growing} it was shown that training cellular automata using neural networks can organize cells into shapes of complex structure. 

\section{Conclusion}
The key technical point of our work is that a complicated Jacobian in the Boyd iteration can be replaced without sacrificing the fooling rate by a single convolution, leading to a remarkably simple low-parametric cellular automata that generates universal adversarial perturbations. It also explains the similarity between UAPs and Turing patterns. It means that the non-linearities are effectively averaged out from the process. This idea could be potentially useful for other problems involving the Jacobians of the loss functions for complicated neural networks, e.g. spectral normalization. 

This work studies theory of interesting phenomena of Universal Adversarial Perturbations. Fundamental susceptibility of deep CNNs to small alternating patterns, that are easily generated, added to images was shown in this work and their connection to Turing patterns, a rigorous theory in mathematical biology that describes many natural patterns. This might help to understand the theory of adversarial perturbations, and in the future to robustly defend from them.

\section*{Acknowlegements}
This work was supported by RBRF grant 18-31-20069.
\bibliography{main}
\include{appendix}

\end{document}

%% file: appendix.tex
\newpage

\section*{Appendix}

\section{Numerical justification of assumptions}

\begin{figure*}[t]
\centering
      \includegraphics[trim=0 360 0 0,clip,width=\textwidth]{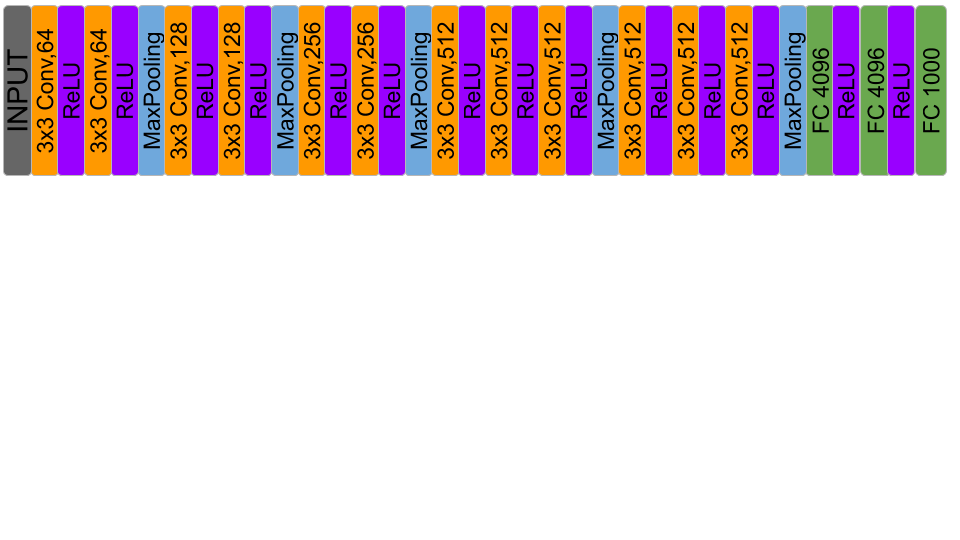}
    \caption{The architecture of VGG-19 \cite{simonyan2014very}}
    \label{arch}
\end{figure*}

To experimentally bolster our assumptions we made following experiments:
\begin{itemize}
    \item Assumption 0
    
    To justify equation \ref{eye} we calculated the average value and the standard deviations of diagonal matrices $D_i$ corresponding to ReLU layers (see Figure \ref{arch}) results are shown in Table \ref{e1}. We assume that Assumption 0 holds, by showing small coefficients of variations for mean values of diagonal elements of the matrices. 
    
    \begin{table}[h!]
\centering
\begin{tabular}{|c|c|c|}\hline
Matrix ($D_i$) & Mean value $E[D_i]$  & Variation (std/mean)\\
\hline
\hline
$D_1$ &  $0.6530 \pm 0.0621$ & $9.5\%$\\
\hline
$D_2$ & $0.4267\pm 0.0362$& $8.48\%$\\
\hline
$D_3$ & $0.4288 \pm 0.0663$ & $15.46\%$\\
\hline
$D_4$ & $0.309\pm0.0401$& $12.97\%$\\
\hline
$D_5$ & $0.3552\pm 0.0515$& $14.49\%$\\
\hline
$D_6$ & $0.3316 \pm0.0578$ & $0.1743\%$\\
\hline
\end{tabular}
\caption{Assumption 0 experiments}\smallskip
\label{e1}
\end{table}

    \item Assumption 1
    We also conduct some experiments to check the Assumption 1, the most correlated elements are the corresponding elements of matrices $D_i \text{ and } D_j$, corresponding to $i$-th and $j$-th ReLU layers (see Figure \ref{arch}). To experimentally test this, we considered two binary vectors (diagonals of matrices). We computed Simple Matching Coefficient (SMC, number of bit matches) between these vectors for each of an image, and averaged over the whole validation dataset.  ISMC shows what fraction of bits are identical in two binary vectors. SMC=1 means total correlation. SMC=0 means total negative correlation, while SMC=0.5 means two vectors are uncorrelated and random to each other.  The results are shown in Table \ref{e2}.
    
    \begin{table}[h!]
\centering
\begin{tabular}{|c|c|c|}\hline
Matrices  & $E[SMC(D_i, D_j)]$\\
\hline
\hline
$D_1$ and $D_2$ & 0.4825 \\
\hline
$D_3$ and $D_4$ & 0.5183\\
\hline
$D_5$ and $D_6$ &0.5528 \\
\hline
$D_6$ and $D_7$ &0.5648 \\
\hline
$D_7$ and $D_8$ &0.6096  \\
\hline
\end{tabular}
\caption{Assumption 1 experiments}\smallskip
\label{e2}
\end{table}

\end{itemize}

\section{Alternative methods of CA optimization}

\begin{table}[t]
\centering
\begin{tabular}{|l|c|c|c|c|}\hline
\diaghead{\theadfont Diag Channel mixing}%
{Target \\ model}{Channel \\ mixing} & \thead{Independent} & \thead{Summation} & \thead{3D filter} & \thead{ Pointwise} \\
\hline
MobileNetV2 & 60.87 & \textbf{94.78} & 81.97 & 94.26 \\
\hline
VGG-19 & 60.74 & 77.50 & \textbf{78.87} & 78.00 \\
\hline
InceptionV3 & 47.45 & 53.10 & 51.31 & \textbf{53.33} \\
\hline
\end{tabular}
\caption{Fooling rates for different channel mixing approaches}\smallskip
\label{tab:mixing}
\end{table}

\begin{figure*}[t]
    \centering
    
    \includegraphics[width=\textwidth]{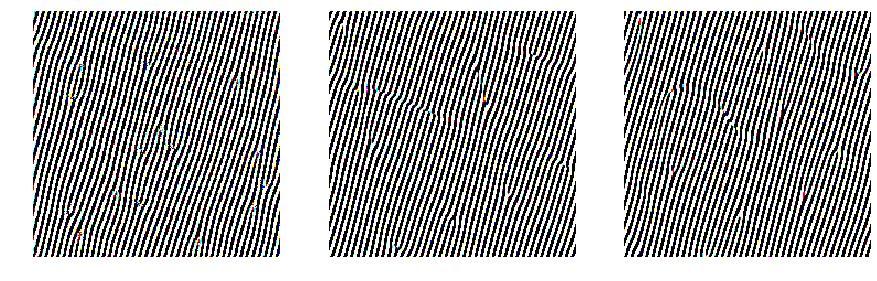}
    
    \caption{Pointwise channel mixing patterns attacking InceptionV3 after Fourier modification }
    \label{fig:fourier_pw}
\end{figure*}

\begin{figure*}[t]
    \centering
    \includegraphics[width=\textwidth]{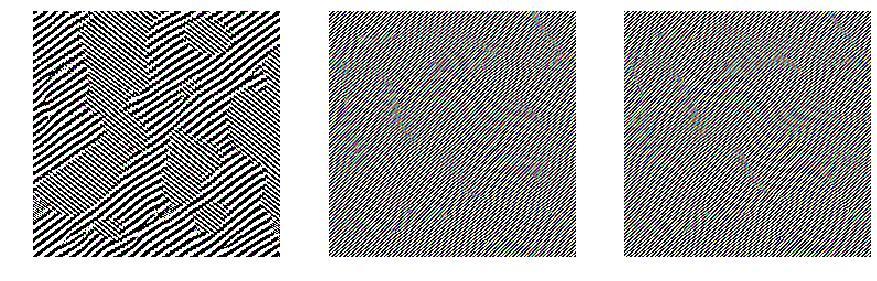}
    \caption{Summation channel mixing patterns attacking InceptionV3 after Fourier modification }
    \label{fig:fourier_sum}
\end{figure*}

\textbf{Effect of different filter sizes.} \label{FS}
Another experiment is the dependence of the fooling rate on the filter size. Here we show results for both patterns with and without specifically optimized initialization maps (attacking MobileNetV2). In both cases small filter size does not provide good results, however surprisingly there is a non-monotonic dependence on the filter sizes once they are larger than $7$. 

\textbf{Effect of channel mixing.} \label{CM}
Different channel mixing strategies within the pattern generation algorithm are used. We compared the following approaches to mixing channel maps during pattern generation:
\begin{itemize}
    \item One 2D-filter for all 3 independent image channels, no channel mixing
    \item One 2D-filter for all image channels, after convolution the sum of each two channels becomes the remaining channel (i.e. \texttt{channel3 = channel1 + channel2})
    \item One 3D-filter for all 3 image channels
    \item 3 separate 2D-filters (one for each channel), and a $3 \times 3$ matrix for channel mixing, like pointwise convolution
\end{itemize}
One can expect for 3D-filter to be a generalization of all of the other cases, thus it would be the most optimal to optimize over. However, as can be seen in Table \ref{tab:mixing}, the results suggest that it is not definitevely better than the other approaches. It can be seen summation, 3D and pointwise approaches give similar results, with pointwise performing slightly better on average over the other models.

\begin{table}
\label{tab:transfer}
    \begin{minipage}[t]{\textwidth}
        \centering
        \begin{tabular}{|l|c|c|c|c|}\hline
        \diaghead{\theadfont Diag Trained onn}%
        {Trained on}{Target} & \thead{MobileNetV2} & \thead{VGG-19} & \thead{InceptionV3} \\
        \hline
        MobileNetV2	&	56.15	&	56.96	&	45.16	\\	\hline
        VGG-19	&	50.49	&	50.06	&	49.98	\\	\hline
        InceptionV3	&	49.55	&	48.76	&	47.18	\\	\hline
        \end{tabular}
        \caption{Turing patterns optimized in $l_1, l_2$}
    \end{minipage}\vspace{0.3cm}
    \begin{minipage}[t]{\textwidth}
        \centering
        \begin{tabular}{|l|c|c|c|c|}\hline
        \diaghead{\theadfont Diag Trained onn}%
        {Trained on}{Target} & \thead{MobileNetV2} & \thead{VGG-19} & \thead{InceptionV3} \\
        \hline
        MobileNetV2	&	90.59	&	61.27	&	35.15	\\	\hline
        VGG-19	&	62.15	&	75.64	&	31.91	\\	\hline
        InceptionV3	&	65.69	&	76.40	&	53.93	\\	\hline
        \end{tabular}
        \caption{Patterns optimized over filter $Y$ and initialization maps}
    \end{minipage}\vspace{0.3cm}
    \begin{minipage}[t]{\textwidth}
        \centering
        \begin{tabular}{|l|c|c|c|c|}\hline
        \diaghead{\theadfont Diag Trained onn}%
        {Trained on}{Target} & \thead{MobileNetV2} & \thead{VGG-19} & \thead{InceptionV3} \\
        \hline
        MobileNetV2	&	60.87	&	45.00	&	37.49	\\
        \hline
        VGG-19	&	61.99	&	60.74	&	47.76	\\
        \hline
        InceptionV3	&	53.22	&	54.54	&	47.45	\\
        \hline
        \end{tabular}
        \caption{Independent channels}
    \end{minipage}\vspace{0.3cm}
    \begin{minipage}[t]{\textwidth}
      \centering
        \begin{tabular}{|l|c|c|c|c|}\hline
        \diaghead{\theadfont Diag Trained onn}%
        {Trained on}{Target} & \thead{MobileNetV2} & \thead{VGG-19} & \thead{InceptionV3} \\
        \hline
        MobileNetV2	&	94.78	&	57.90	&	35.04	\\	\hline
        VGG-19	&	73.15	&	77.50	&	43.06	\\	\hline
        InceptionV3	&	63.90	&	64.46	&	53.10	\\	\hline
        \end{tabular}
    \caption{Summation channel mixing}
    \end{minipage}\vspace{0.3cm}
    \begin{minipage}[t]{\textwidth}
      \centering
        \begin{tabular}{|l|c|c|c|c|}\hline
        \diaghead{\theadfont Diag Trained onn}%
        {Trained on}{Target} & \thead{MobileNetV2} & \thead{VGG-19} & \thead{InceptionV3} \\
        \hline
        MobileNetV2	&	81.97	&	68.30	&	48.24	\\	\hline
        VGG-19	&	76.15	&	78.87	&	43.83	\\	\hline
        InceptionV3	&	61.36	&	61.62	&	51.31	\\	\hline
        \end{tabular}
    \caption{3D filter}
    \end{minipage}\vspace{0.3cm}
    \begin{minipage}[t]{\textwidth}
      \centering
        \begin{tabular}{|l|c|c|c|c|}\hline
        \diaghead{\theadfont Diag Trained onn}%
        {Trained on}{Target} & \thead{MobileNetV2} & \thead{VGG-19} & \thead{InceptionV3} \\
        \hline
        MobileNetV2	&	94.26	&	57.80	&	35.05	\\	\hline
        VGG-19	&	74.79	&	78.00	&	43.38	\\	\hline
        InceptionV3	&	59.29	&	67.70	&	53.33	\\	\hline
        \end{tabular}
    \caption{Pointwise channel mixing}
    \end{minipage}    
\end{table}

\subsection{Transferability}
We tested several patterns discussed earlier for their transferability among the target networks. The results can be seen in Table \ref{tab:transfer}. It seems that the attacks generally transfer rather well to MobileNetV2 anf VGG-19, while InceptionV3 seems harder to transfer this types of attacks to. The results show that while optimization over cellular automata parameters seems to significantly boost fooling rates for target networks, some transferability loss can be expected. Note that for MobileNetV2 we get almost $95\%$ fooling rate.

\begin{figure*}[t]
\centering
    \begin{subfigure}{.49\textwidth}
      \centering
      \includegraphics[width=\textwidth]{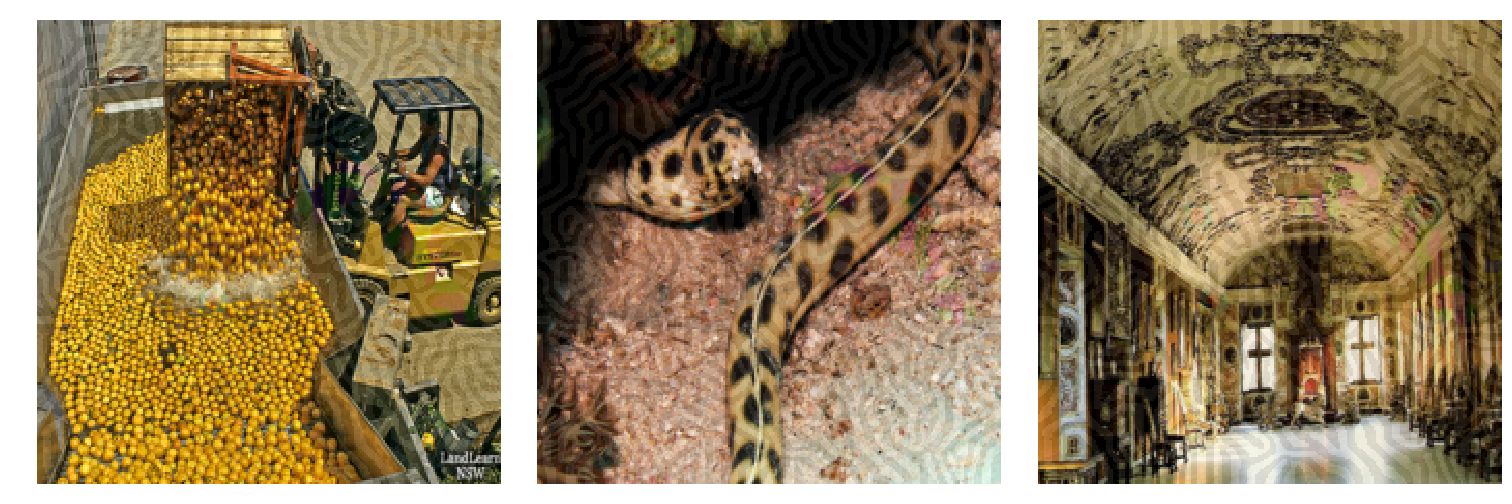}
  \caption{Examples of images misclassified after the perturbation with UAPs from \cite{khrulkov2018art}($\epsilon=10$)}
    \end{subfigure}\hfill
    \begin{subfigure}{.49\textwidth}
      \centering
      \includegraphics[width=\textwidth]{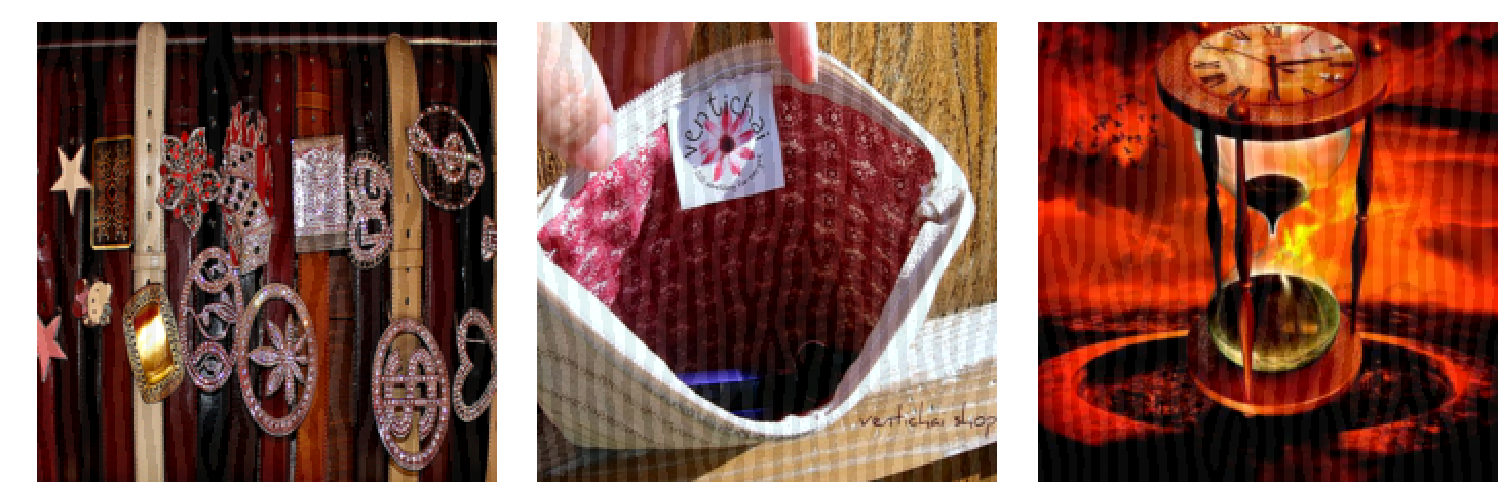}
  \caption{Examples of images misclassified after the perturbation with Turing patterns ($\epsilon=10$).}
      \label{fig:turing_true_vgg}
    \end{subfigure}\hfill
  \label{fig:fig2}
  \caption{Examples of images misclassified after the perturbation with UAPs and Turing patterns.}
\end{figure*}

\begin{algorithm}
\SetAlgoLined
 initialize $<x>$\\
 $\boldsymbol{C}\leftarrow \boldsymbol{I}$;\\
 \Repeat{Termination criterion fulfilled}{
  $\boldsymbol{B}, \boldsymbol{D} \leftarrow eigendecomposition(\boldsymbol{C}$)\\
  \For{$i = 1 \leftarrow \lambda$}{
   $y_i \leftarrow \boldsymbol{B} \cdot \boldsymbol{D} \cdot \mathcal{N}(\boldsymbol{0}, \boldsymbol{I})$\\
   $x_i \leftarrow <x> + \sigma y_i$\\
   $f_i \leftarrow f(x_i)$\\
  }
  $<y> \leftarrow \sum_{i=1}^{\mu} w_i y_{i:\lambda}$\\
  $<x> \leftarrow <x> + \sigma <y> = \sum_{i=1}^{\mu} w_i x_{i:\lambda}$\\
  $\sigma \leftarrow update(\sigma, \boldsymbol{C}, y)$\\
  $\boldsymbol{C} \leftarrow update(\boldsymbol{C}, w, y)$\\
 }
 \caption{($\mu_{w}, \lambda$)-CMA-ES algorithm}
\end{algorithm}